\documentclass[sn-mathphys]{sn-jnl}

\usepackage{lineno,hyperref}
\usepackage{underscore}

\jyear{2021}%

\theoremstyle{thmstyleone}%
\theoremstyle{thmstyletwo}%
\theoremstyle{thmstylethree}%

\begin{document}

\title[DRA-OVG model for Vehicle Re-identification]{Discriminative-Region Attention and Orthogonal-View Generation Model for Vehicle Re-Identification}


\author[1,3]{\fnm{First} \sur{Huadong Li}}

\author[1,2]{\fnm{Second} \sur{Yuefeng Wang}}

\author*[1,2]{\fnm{Third} \sur{Ying Wei}}\email{weiying@ise.neu.edu.cn}
\author[1]{\fnm{Fourth} \sur{Lin Wang}}
\author[3,4]{\fnm{Fifth} \sur{Li Ge}}

\affil*[1]{\orgdiv{College of Information Science and Engineeringt}, \orgname{Northeastern University}, \orgaddress{\street{Nanhu}, \city{Shenyang}, \postcode{110819}, \state{Liaoning}, \country{China}}}

\affil[2]{\orgdiv{Information Technology R\&D Innovation Center of Peking University}, \orgname{Peking University}, \orgaddress{\street{Gaobu}, \city{Shaoxing}, \postcode{312000}, \state{Zhejiang}, \country{China}}}

\affil[3]{\orgdiv{Peng Cheng Laboratory}, \orgaddress{\street{Xingke 1st}, \city{Shenzhen}, \postcode{518000}, \state{Guangdong}, \country{China}}}
\affil[4]{\orgdiv{School of Electronics and Computer Engineering},\orgname{Peking University Shenzhen Graduate School}, \orgaddress{\street{Lishui}, \city{Shenzhen}, \postcode{518055}, \state{Guangdong}, \country{China}}}

\abstract{Vehicle re-identiﬁcation (Re-ID) is urgently demanded to alleviate the pressure caused by the increasingly onerous task of urban traffic management. Multiple challenges hamper the applications of vision-based vehicle Re-ID methods: (1) The appearances of different vehicles of the same brand/model are often similar; However, (2) the appearances of the same vehicle differ significantly from different viewpoints. Previous methods mainly use manually annotated multi-attribute datasets to assist the network in getting detailed cues and in inferencing multi-view to improve the vehicle Re-ID performance. However, finely labeled vehicle datasets are usually unattainable in real application scenarios. Hence, we propose a Discriminative-Region Attention and Orthogonal-View Generation (DRA-OVG) model, which only requires identity (ID) labels to conquer the multiple challenges of vehicle Re-ID. The proposed DRA model can automatically extract the discriminative region features, which can distinguish similar vehicles. And the OVG model can generate multi-view features based on the input view features to reduce the impact of viewpoint mismatches. Finally, the distance between vehicle appearances is presented by the discriminative region features and multi-view features together. Therefore, the significance of pairwise distance measure between vehicles is enhanced in a complete feature space. Extensive experiments substantiate the effectiveness of each proposed ingredient, and experimental results indicate that our approach achieves remarkable improvements over the state-of-the-art vehicle Re-ID methods on VehicleID and VeRi-776 datasets.}




\maketitle
\section{Introduction}\label{sec1}
The purpose of vehicle Re-ID is to retrieve the images of a target vehicle from images taken by multiple traffic surveillance cameras\cite{zapletal2016vehicle}. Vehicle Re-ID has been widely studied because of its potential applications in ancillary traffic management and intelligent surveillance. Nevertheless, due to the unique 3-D structure and the standardized production mode, the vehicle Re-ID task is more challenging than the similar problem called person Re-ID\cite{borgia2018cross,chen2016deep,zheng2019joint,yan2019learning,yu2019unsupervised,yu2018unsupervised,li2017person,xiang2018person}, which has achieved outstanding outcomes.

\begin{figure}[htb]
	\center{
		\includegraphics[width=8cm]{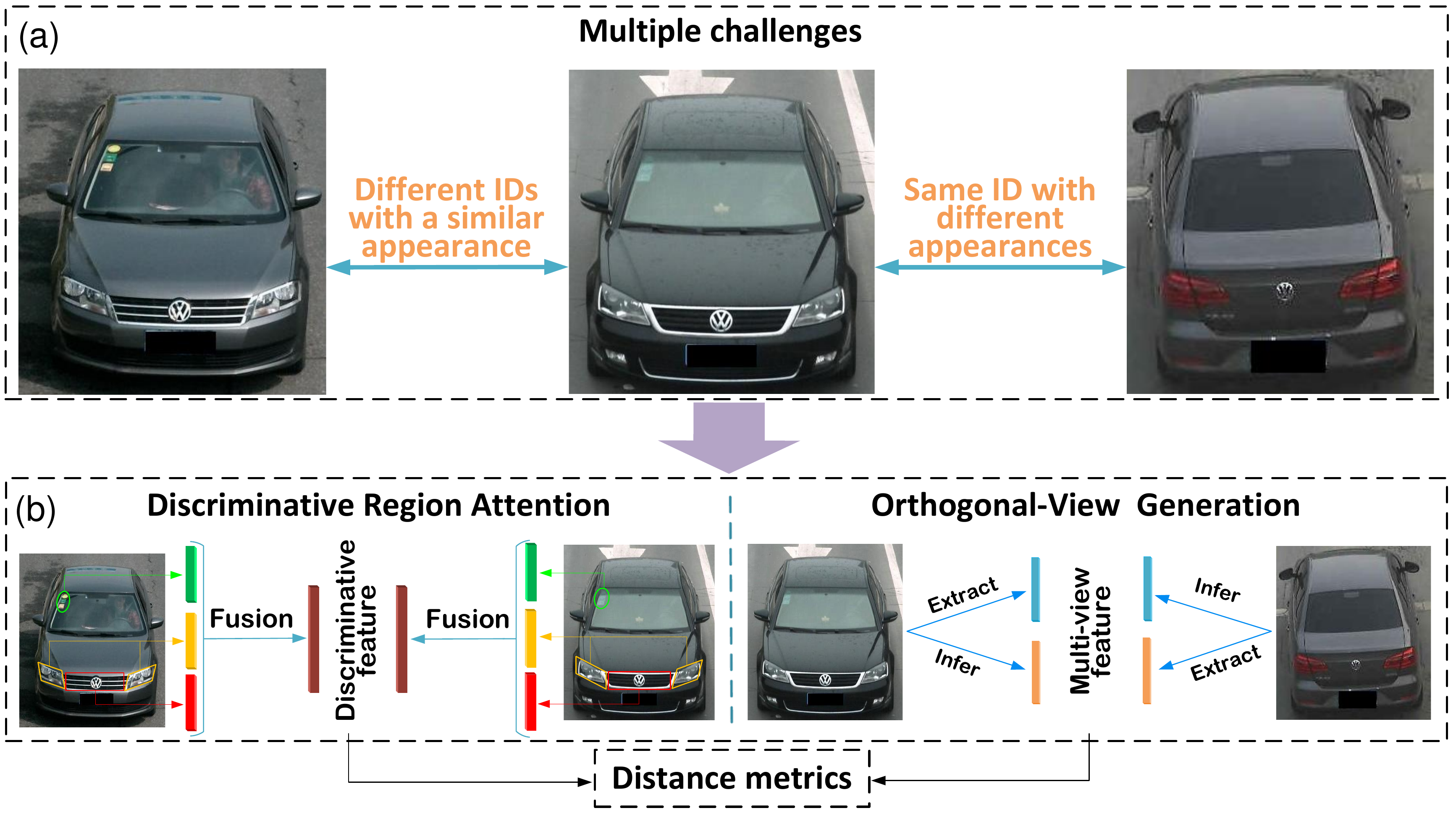}
		\caption{(a) The main challenges of vehicle Re-ID. (b) The basic intention of our method.
	} \label{picture one}}
\end{figure}

As shown in Fig. \ref{picture one} (a), from the same viewpoint, the appearances of different vehicles may be highly similar. On the other hand, various viewpoints often lead to false-negative cases. Especially the viewpoints have no overlapping fields, which are almost uncorrelated in the feature space, such as the front and back of vehicles (the two flanks of a vehicle are always symmetrical and similar). Thus,  we define the front and back views as a pair of orthogonal views.

To distinguish different vehicles with similar appearances, some recently proposed vehicle Re-ID methods attempt to make full use of discriminating details \cite{liu2018ram,wang2018learning}. However, the detail-based methods suffer some disadvantages such as incorrect localization of details and the need for a large amount of annotation data \cite{zhao2019structural, guo2019two, lou2019embedding}, limiting their application in practical scenarios.

In order to solve the viewpoint mismatch problem when comparing vehicle images, scholars proposed some multi-view inference models \cite{zhou2018aware}. Recently, the popular Generative Adversarial Network (GAN) has been introduced to transform the original view features into multi-view features \cite{goodfellow2014generative}. Research \cite{zhou2018vehicle} involves manually marked viewpoint labels instead of the traditional min-max games to optimize the network since the transformation of features for vehicle Re-ID has a specific direction. However, viewpoint information can hardly be acquired due to the high label cost in real-world scenarios.

Analyzing the existing datasets shows that vehicle ID is a compulsory label and is relatively easy to obtain in natural scenes. Therefore, we propose a discriminative-region attention and orthogonal-view generation model (DRA-OVG), which only utilizes ID labels to construct the identity feature of vehicles in a complete feature space. Fig. \ref{picture one} (b) shows the basic strategy of our method: Given a pair of images, under unsupervised conditions, the DRA model extracts their discriminative region features, and the OVG model generates their orthogonal-view features from the original-view features. Finally, the discriminative and multi-view features are combined to optimize the distance metrics.

The main contributions of this paper are highlighted as follows:

(1) We propose a prototype generation module that can generate semantic prototypes by leveraging the mapping rule in feature space beneath an image set. These prototypes can automatically locate different semantic regions of vehicle images. Thereby, the algorithm's dependence on fine-grained labeled data is greatly reduced.

(2) A discriminative region attention model (DRA) is proposed to address the high intra-class similarity problem. The DRA model uses semantic prototypes to extract discriminative region features containing more vehicle details, which help encode vehicle identities to distinguish between similar vehicles.

(3) We propose an orthogonal-view generation (OVG) model to address the problem of high intra-class discrepancy. In the OVG model, we combine the features of two special local semantics to construct the vehicle viewpoint feature, which can deduce a vehicle's viewpoint. We use the viewpoint feature to automatically extract the supervisory information for training the generation network to realize orthogonal-view generation without additional manual annotation. Then, we align vehicle features in the viewpoint domain to reduce the negative impact of viewpoint differences. 

The rest of the paper is organized as follows: Section II reviews some related works. The proposed discriminative region attention and orthogonal-view generation model are presented in Section III. The proposed methods are evaluated on VehicleID and VeRi-776 datasets and are compared with several existing state-of-the-art methods in Section IV. Section V discusses the advantages and limitations of our method, and Section VI concludes this paper.

\section{RELATED WORKS}
\subsection{Re-ID by Traditional Method and DNN}
 With the extensive use of deep neural networks (DNN), more and more vehicle Re-ID methods use DNN to extract vehicle features for making further reformations. Sun et al. \cite{sun2018visual} fused vehicle features extracted by ResNet50 \cite{he2016deep} and GoogLeNet \cite{yang2015large} to encode vehicle images for getting powerful descriptors. In addition to using the existing networks, some people designed new network structures for vehicle Re-ID. For example, Zhu et al. \cite{zhu2019vrsdnet} proposed a shortly and densely connected convolutional neural network (SDC-CNN) for vehicle Re-ID. The special structure can significantly enhance the feature learning ability of the network. With vehicle orientation and metadata attributes being considered, Huang et al.\cite{huang2019multi} proposed a viewpoint-aware temporal attention model for vehicle Re-ID by utilizing deep-learning features extracted from consecutive frames. 

\subsection{Fine-Grained Re-ID}
 In the recognition field, some traditional methods have yielded outstanding results. For example, \cite{2021Gesture} proposed a slope difference distribution method (SDD) to define and extract shape features for gesture recognition. SDD can effectively solve the problem of object recognition with different shapes, such as gesture recognition. However, it is not suitable for vehicle Re-ID. Because vehicles usually have very similar shapes, the difference between them only exists in the detailed appearance features. Thus, the slope feature cannot be used to identify vehicles. With the profound understanding of vehicle Re-ID, numerous emerging works tried to improve Re-ID's performance by using subtle visual differences. Considering the specific structures of vehicle images, Liu et al. \cite{liu2018ram} proposed a region-aware deep model (RAM) to extract features from a series of local regions by hard segmentation and separately trained each branch to increase the network’s attention to details. Guo et al. \cite{guo2019two} proposed a two-level attention network model. The model is composed of a hard part attention module and a soft pixel attention module, which can adaptively extract the discriminatory features from the visual appearance of vehicles. Khorramshahi et al.\cite{khorramshahi2019attention} leveraged an attention-based model, which could learn to focus on different parts of a vehicle by conditioning the feature maps on visible key points. In order to utilize the relationship between visual appearances under different levels. Wei et al. \cite{wei2018coarse} proposed a segmentation end-to-end RNN-based hierarchical attention (RNN-HA) classification model for vehicle Re-ID. The RNN-HA model consists of three coupled modules that capture different levels of vehicle appearances to describe a vehicle comprehensively. For a better combination of detail cues and global appearances, He et al. \cite{he2019part} proposed a part-regularized discriminatory-feature preserving method to enhance the perceptive ability of subtle discrepancies and developed a novel framework to integrate the part constraints with the global Re-ID models by introducing a detection branch. Furthermore, Wang et al.\cite{wang2020attribute} proposed a novel attribute-guided network (AGNet), which could learn global representation with abundant attribute features in an end-to-end manner. Specifically, an attribute-guided module is proposed in AGNet to generate an attribute mask, which could inversely guide the selection of discriminative features for category classification. He et al.\cite{2021TransReID} proposed a pure transformer-based Re-ID method named TransReID. TransReID introduces side information embeddings (SIE) to mitigate feature bias towards camera/view variations by plugging in learnable embeddings to incorporate these non-visual clues. Specifically, the method encodes the camera and viewpoint labels into 1-D embeddings. Then, the embeddings are fused with visual features as positional embeddings to address the feature bias towards the camera/view variations problem. This method achieves state-of-the-art performance on both person Re-ID and vehicle Re-ID tasks. Quispe  et al.\cite{quispe2021attributenet} proposed AttributeNet (ANet) that jointly extracts identity-relevant features and attribute features. ANet enables interaction by distilling the Re-ID helpful attribute feature and adding it into the general Re-ID feature to increase discrimination. However, all of these fine-grained methods require additional attribute annotations for helping them to get fine-grained features extraction ability, limiting their applicability.

 \subsection{Generation Based Re-ID Model}
 Enormous methods were devised to solve the problem of multi-view differences. Zhou et al. \cite{zhou2018vehicle} proposed two end-to-end depth structures: The spatially concatenated ConvNet and the CNN-LSTM bi-directional loop. The two structures take advantage of CNN and long short-term memory (LSTM) to learn the transformations across different viewpoints of vehicles. Due to GAN’s\cite{goodfellow2014generative} advantages in image generation, some scholars also used the GAN to generate multi-view features of vehicles. Zhou et al. \cite{zhou2018aware} proposed a view-aware attention multi-view inference (VAMI) model. Given a vehicle image from an arbitrary viewpoint, VAMI infers the multi-view features through GAN to optimize pairwise distance metrics’ learning. Wang et al. \cite{wang2020kernelized} proposed a general framework named kernelized multi-view subspace analysis (KMSA) for multi-view data dimension reduction. KMSA directly handles the multi-view feature representation in the kernel space, which provides a feasible channel for direct manipulations of multi-view data with different dimensions. Besides, to address the severe domain bias problem, Peng et al.\cite{peng2020cross} proposed a domain adaptation framework for vehicle Re-ID (DAVR), which narrows the cross-domain bias by fully exploiting the labeled data from the source domain to adapt to the target domain. 
 \begin{table}[]
	\centering
	\caption{Notations used in the paper.} \label{Table one}
	\begin{tabular}{c | l }
		\hline
		Notation & Description \\ \hline\hline
		$x$    & one feature in feature maps      \\
		$X_{set}$  &  a set of all features of the vehicles in the subset  \\
		$I$    & one image     \\
		$l$    & the relationship between two images \\
		$r$    & the regional semantic of vehicle images     \\
		$d_r$  & the prototype of the semantic r     \\	
		$L_r$  & the location indication matrix of the semantic r   \\
		$p^r$  & the probability of the occurrence of semantic r   \\
		$P^r$  & the probability matrix of semantic r\\	
		$f_g$  & the generated orthogonal-view feature  \\
		$\hat{f_g}$  & the real orthogonal-view feature  \\
		$f_d$  & the fused discriminative region feature     \\
		$f_o$  & the original view feature     \\
		$f_v$  & the viewpoint feature     \\
		$\mathcal L $  &loss function \\ \hline

	\end{tabular}

\end{table}
\section{Proposed Methods}
\subsection{Preliminaries and Problem Formulation}
Common notes, shown in table \ref{Table one}, are used throughout the paper. Given an image $I$ of size $H\times W\times C$, we use the feature extraction network $F$ to extract base features, and the activations of a middle layer in $F$ are called "feature maps" with $h\times w \times c$ elements. We consider the feature maps as having $h\times w$ positions with each position containing a c-dimensional feature $x$. The corresponding local semantic of the original image is encoded in $x$ \cite{zeiler2014visualizing}. We take the features of an image set to construct a feature set $X_{set}$, namely $X_{set}=\{x_{ij}^n$ $\in$ $R^c\}$, where $x_{ij}^n$ represents the feature at the position $(i,j)$ ($i\in \{1,\cdots,h\}$, $j\in \{1,\cdots,w\}$) in the feature maps of image $n$.

A pair of images are defined as $(I^a, I^b)$, and $l^{ab}$ is the identity relationship between them. If $I^a$ and $I^b$ belong to the same vehicle, $l^{ab}=1$, otherwise, $l^{ab}=0$. For an image $I$, we aim to get its discriminative feature and generate its orthogonal-view feature by the following functions:
\begin{equation}
\setlength{\abovedisplayskip}{9pt}
\setlength{\belowdisplayskip}{0.5pt}
f_d=T(F(I), \{L_r\}_{r=1}^R),
\end{equation}

\begin{equation}
\setlength{\abovedisplayskip}{0.5pt}
\setlength{\belowdisplayskip}{3pt}
f_g=G_f(F(I)),
\end{equation}
where $F(\cdot)$ is the feature extraction network, and $G_f(\cdot)$ generates orthogonal-view features base on the original view. $T(\cdot)$ fuses features of different semantics. $L_r$ is the indication matrix that indicates the location of the specific semantic $r$, and $R$ represents the defined number of discriminative regions. $f_g$ and $f_d$ represent the generated orthogonal-view feature and the fused discriminative region feature, respectively. After defining $f_g$ and $f_d$, we aim to design and optimize $T(\cdot)$, $G_f(\cdot)$, and $L_r$ to shorten the distance between $f_g^a$ and $f_g^b$, $f_d^a$ and $f_d^b$ when $l^{ab}=1$, and maximize the distance when $l^{ab}=0$ by using the distance metrics loss \cite{schroff2015facenet}.

\subsection{OVERVIEW OF THE METHOD}
As shown in Fig. \ref{picture one} (b), the DRA-OVG model is divided into two parts: a DRA model with advantages of extracting and fusing discriminative region features and an OVG model with advantages of orthogonal-view feature generation.
\begin{figure*}[htb]
	\center{
		\includegraphics[width=12cm]{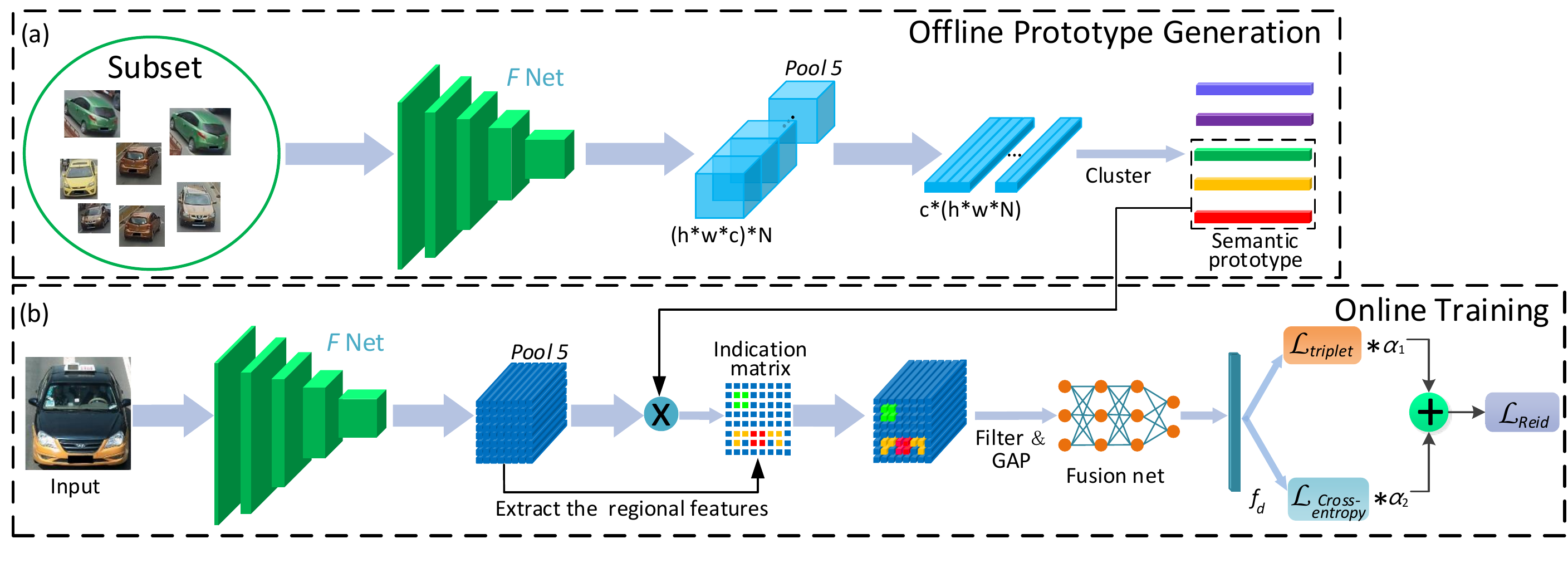}
		\caption{The overview of the framework of DRA. (a) The offline generation process of semantic prototypes and (b) the process of extracting and re-encoding the features of discriminative regions.
	}\label{picture two} }
\end{figure*}

Fig. \ref{picture two} shows the structure of DRA. After encoding the image, we leverage the mapping consistency of the same semantic regions in feature space to generate semantic prototypes offline, which can be used to calculate the indication matrix $L_r$. Then, under the guidance of $L_r$, the features of discriminative regions can be extracted. Finally, the feature fusion network $T(\cdot)$ encodes regional features as discriminative feature $f_d$. Besides, the structure of the OVG model is illustrated in Fig. \ref{picture four}. We design a viewpoint feature generator to construct vehicles' viewpoint features $f_v$. The distribution law of $f_v$ in the high-dimensional feature space is applied to get viewpoint discriminators, which can help us automatically extract real orthogonal-view features $\hat{f_g}$ for each image. We use $\hat{f_g}$ to supervise the learning of $G_f(\cdot)$. The optimized $G_f(\cdot)$ can generate orthogonal-view features $f_g$ based on original view features $f_o$.  During testing, $f_o$, $f_g$, and $f_d$ are combined to express vehicles' identity. The details of each part are described in the following subsections.

\subsection{Discriminative Region Attention}
When distinguishing two similar vehicles, people often focus on the differences between the same semantics. For example, they may compare the appearances of lights or the appearances of air intake grilles of two vehicles. Based on the same idea, we designed the DRA model to deal with the problem of identifying similar vehicles.
\subsubsection{Offline prototypes generation}
Unlike previous methods\cite{liu2018ram, guo2019two}, we do not focus on studying a single image but leverage the mapping consistency beneath an image set to generate semantic prototypes. Fig. \ref{picture two} (a) shows the generation process of the prototypes. First, we use $F$ to extract features of all vehicles in a subset, and all features of the vehicles construct $X_{set}$. At this time, each feature $x_{ij}^n$ encodes a local semantic of the original image, and the features which represent the same semantic should be relatively close in the feature space. Therefore, we use the spectral clustering algorithm to mine the mapping consistency. We cluster the $X_{set}$ into five classes and the center vectors of each class are used as semantic prototypes: $\{d_1,d_2,\cdots,d_5\}$, $d_r\in R^c$. The location of the semantic $r$ is detected by $d_r$. In the DRA model, we only use the prototype generation module to generate prototypes once, and it will not be used in the online training stage.

\subsubsection{Location indication matrix}
The probability of the occurrence of semantic $r$ at each position is expressed as:
\begin{equation}
p_{ij}^r=max(\frac{d_r}{\mid d_r\mid} \cdot \frac{x_{ij}}{\mid x_{ij}\mid},0),
\end{equation}
where $\mid \cdot \mid$ denotes the norm of the vector. According to the meaning of $p_{ij}^r$, all $p_{ij}^r$ of an image construct a probability matrix $P^r$ whose dimensions are $h\times w$:
\begin{equation}
P^r=
\begin{bmatrix}
p_{(1,1)}^r & \cdots & p_{(1,w)}^r \\
\vdots & \ddots & \vdots \\
p_{(h,1)}^r & \cdots & p_{(h,w)}^r
\end{bmatrix}
\in R^{h\times w},
\end{equation}
To facilitate calculation, we convert the probability matrix into a binary form to get the indication matrix, which can help us extract local semantic features of the vehicle by basic matrix operations:
\begin{equation}
L_r=sgn (P^r),
\end{equation}
where $sgn(\cdot)$ is a sign function. The element in $L_r$ takes 1 where its corresponding element in $P^r$ is positive and takes 0 otherwise.
\label{sec:guidelines}

\subsubsection{Discriminative features extraction and fusion}

Feature maps of the $pool5$ layer in $F$ are used to extract discriminative region features. We use $\{d_1,d_2,\cdots,d_5\}$ to generate indication matrices, and then various semantic regions of vehicle images are positioned according to the matrices. As shown in Fig \ref{picture six}, after visualizing the positioning results, we find that the semantic prototypes can detect the regions of annual inspection stickers, lights, air intake grilles, seats and background of vehicles. It is worth noting that $F$ trained for vehicle Re-ID automatically focuses on these semantic meanings by its convolution kernels. We only mine and leverage this rule rather than design it. According to the data in Tabel \ref{Table four}, we select the regions of the annual inspection sticker, light, and intake grille as the discriminative regions, and we present their prototypes as $\{d_1,d_2,d_3\}$. In the online learning stage, $\{d_1,d_2,d_3\}$ are used to extract discriminative region features. Then we concatenate these features for global average pooling. The fusion net is designed as a fully connected network with four layers. As illustrated in Fig. \ref{picture two} (b), classification loss and triplet loss are used together to optimize the fusion network. $\alpha_1$ and $\alpha_2$ adjust the proportional relationship between these loss functions. The whole process of generating $f_d$ is shown in Algorithm \ref{alg:one}.

\begin{algorithm}[htb]
	\caption{Discriminative features fusion}
	\label{alg:one}
	\begin{algorithmic}[1]
		\Require{Input image $I$, prototypes $\{d_1, d_2, d_3\}$};
		\Ensure{Fused discriminative region feature $f_d$};
		
		\State 	Feed $I$ into $F$ to get feature maps of $pool5$;
		\For{$r=1,2,3$}
		\State	Use $d_r$ and the feature maps of $pool5$ to compute $L_r$ by (3), (4), and (5);
		\EndFor
		\State Feed the feature maps of $pool5$, $L_1$, $L_2$, and $L_3$ into (1) to compute $f_d$;
		\State $\mathbf{return}$ $f_d$
	\end{algorithmic}
\end{algorithm}

\begin{figure*}[htb]
	\center{
		\includegraphics[width=12cm]{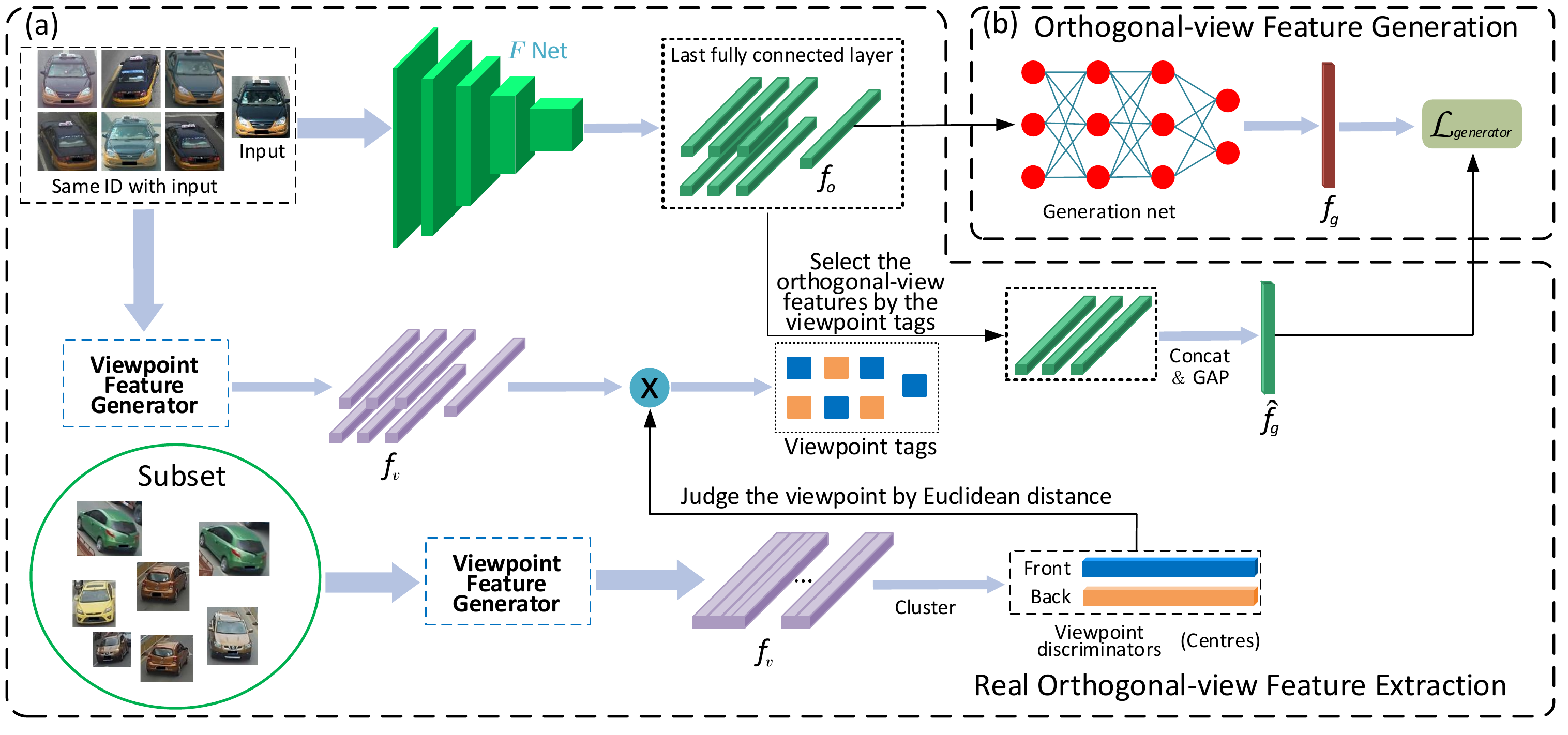}
	}
	\caption{The overview of the structure of OVG. (a) The extraction process of the real orthogonal-view features and (b) the learning process of orthogonal-view feature generation. 
	\label{picture four} }
\end{figure*}
\begin{figure*}[htb]
	\center{
		\includegraphics[width=10cm]{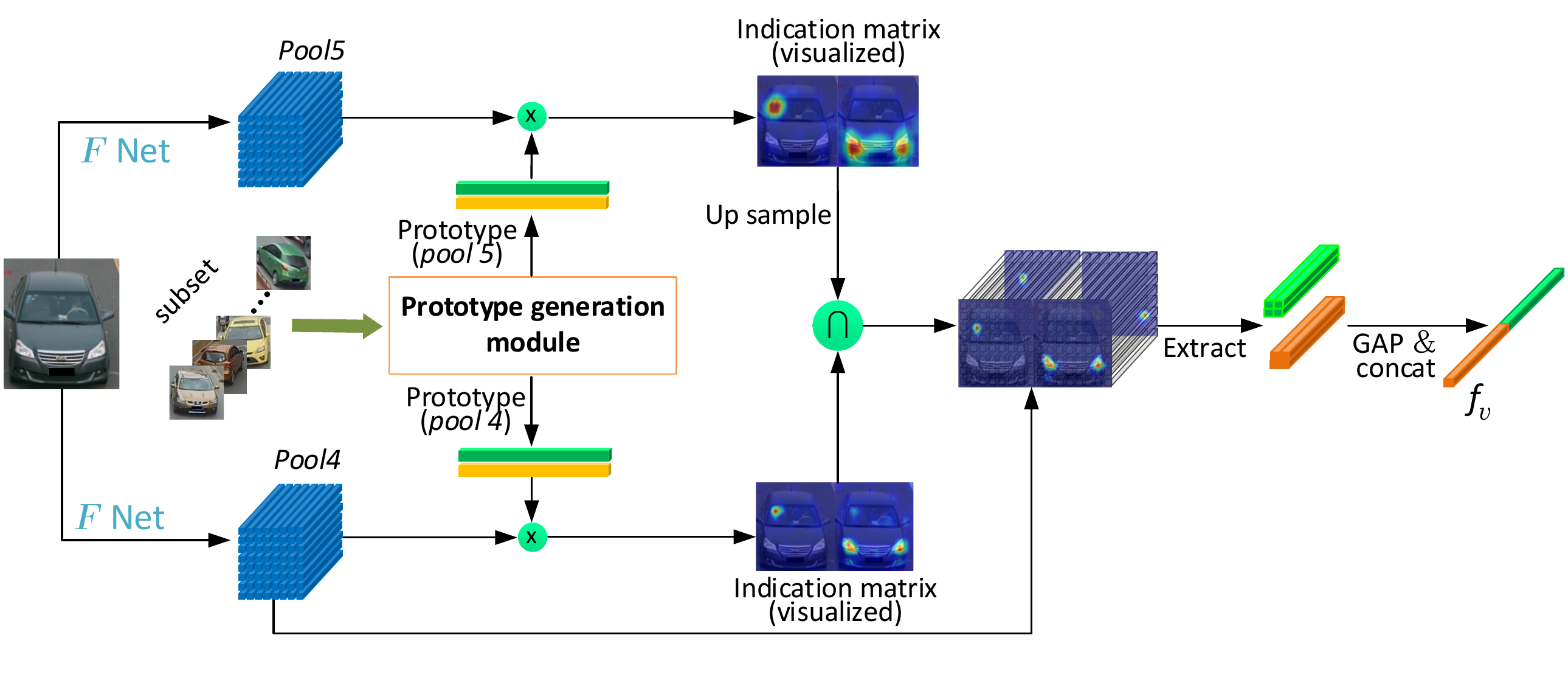}
		\caption{The details of the viewpoint feature generator. The translucent heat maps visually show the internal fundamental of the model.
	} \label{picture three}}
\end{figure*}
\subsection{Orthogonal-View Feature Generation}
One can easily infer the orthogonal view based on a given view of a vehicle. That is because people have learned this corresponding relation from a lot of life experiences. Based on the same idea, we design the OVG model to infer the orthogonal-view feature from the original view feature by learning the corresponding relation between the features of orthogonal-view pairs.

\subsubsection{Viewpoint feature generator}
Generation of vehicles' viewpoint features is the core of our OVG model, so we first introduce the viewpoint feature generator, whose structure is illustrated in Fig. \ref{picture three}. Since the features of lights and annual inspection stickers are noticeably different between orthogonal viewpoints, we use those features to construct viewpoint feature ${f_v}$ for each vehicle.

Identifying whether the annual inspection stickers are present or not and the colors of lights are vulnerable to the surrounding unrelated information. So we hope to locate these local semantic regions more accurately to reduce the influence of irrelevant factors.

Unfortunately, we cannot finely position these semantic regions of vehicles through the feature maps of a single layer. The prototypes generated by the feature maps of $pool5$ are robust to interference, but the location is not precise enough. That is because deep convolution kernels have a large receptive field, so the semantics expressed by deep convolution features are more advanced, and no similar semantics would appear in the same image. However, this also means the semantic regions located by deep layer feature maps are so coarse that they cannot be utilized to identify viewpoints. On the contrary, the prototypes generated by the feature maps of $pool4$ can subtly locate the target semantic regions, but it is fallible. That is also understandable because the receptive field of the shallow convolution kernels is narrow, so their semantics are basic and similar. Therefore, utilizing the intersection of the indication matrices computed by the feature maps of $pool4$ and $pool5$ layers to accurately and robustly locate the regions of vehicles' annual inspection stickers and lights is a better solution. We have also tried to use the feature maps of other layers to locate stickers and lights, but the outcomes are not satisfactory.

As shown in Fig. \ref{picture three}, we put a set of images into $F$ to extract feature maps, then the feature maps of $pool5$ and $pool4$ layers are fed into the prototype generation module, which shares the same structure with that in Fig. \ref{picture two}(a), to generate semantic prototypes $\{d_1^4,d_2^4  ,\cdots,d_5^4 \}$ and $\{d_1^5,d_2^5  ,\cdots,d_5^5 \}$, respectively. The local semantic indication matrix is calculated by (3), (4), and (5). The probability matrices $P^r$ are up-sampled to the same size as the input image by bilinear interpolation to generate heat maps. As shown in Fig. \ref{picture five}, we intuitively select the annual inspection sticker prototype and the light prototype, represented as $\{d_1^4,d_2^4 \}$ and $\{d_1^5,d_2^5 \}$, by those heat maps.

After obtaining the prototypes, the viewpoint feature generator can generate an image's viewpoint feature easily, as detailed in algorithm \ref{alg:two}.

\begin{algorithm}[htb]
	\caption{{Viewpoint feature generation}}
	\label{alg:two}
	\begin{algorithmic}[1]
		\Require{{Input image $I$, $\{d_1^4,d_2^4\}$, $\{d_1^5,d_2^5\}$ and $F$;}}
		\Ensure{{Viewpoint feature $f_v$ of $I$;}}
		
		\State 	{Feed $I$ into $F$ to get feature maps of $pool4$ and $pool5$;}
		\For{{$r=1,2$}}
		\State	{ Use $d_r^4$ and the feature maps of $pool4$ to compute $L_r^{4}$ by (3), (4), and (5);}
		\State	{ Use $d_r^5$ and the feature maps of $pool5$ to compute $L_r^{5}$ by (3), (4), and (5);}
		\State  { Up-sample $L_r^5$ to the size of $L_r^4$;}
		\If{{ r=1}}
		\State { $L_r^{inter}\gets L_r^4\cap L_r^5$;}
		\State { Extract $f_r$ from feature maps of $pool4$ according $L_r^{inter}$;}
		\State { $f_v\gets GAP(f_r)$;}
		\Else
		\State { $L_r^{inter}\gets L_r^4\cap L_r^5$;}
		\State {Extract $f_r$ from feature maps of $pool4$ according $L_r^{inter}$;}
		\State {$f_v\gets concat(f_v,GAP(f_r))$;}
		\EndIf
		\EndFor
		\State {$\mathbf{return}$ $f_v$.}
	\end{algorithmic}
\end{algorithm}

\subsubsection{Real orthogonal-view feature extraction}
With only the vehicle ID label, we hope to design an algorithm that automatically extracts supervisory information for $G_f(\cdot)$ to reduce the cost of manual annotation. Fig. \ref{picture four}(a) demonstrates the process intuitively. First, we use the viewpoint feature generator model to generate the viewpoint features of all vehicles in the subset, and then we cluster these features into two clusters. We take the center vectors of the two clusters as the discriminators of the two viewpoints. 
Since the annual inspection sticker and light only appear simultaneously in the front view, the images containing the vehicle's front view are grouped into one group. 
The remaining images (mainly containing the back views of vehicles) are divided into the other group. Therefore, we define the centers of these two groups as the front-view and back-view discriminator, respectively. Given a certain image, all images that share the same ID in the training set can be found according to the ID label. We put these images into the viewpoint feature generator to generate viewpoint features and then calculate their distance between the viewpoint discriminators to judge the viewpoints of the images. The view features orthogonal to the original view are selected according to the viewpoint tags representing each image's viewpoint. Finally, the real orthogonal-view feature $\hat{f_g}$ is obtained by global average pooling.

\subsubsection{Orthogonal-view feature generation}
We design $G_f(\cdot)$  as a four-layer fully connected network, which is shown in Fig. \ref{picture four} (b). It receives the global feature of the original view and outputs the orthogonal-view feature. The loss function is formalized as follows:
\begin{equation}
\mathcal{L}_{generator}=\min_{\substack{G_f}}\sqrt{(G_f(f_o)-\hat{f_g})^2},
\end{equation}
where $f_o$ is the output of the last fully connected layer of $F$ and $\hat{f_g}$ is the real orthogonal-view feature of the input image. In the testing phase, we use $G_f(\cdot)$ to generate orthogonal-view features of vehicles, and the real orthogonal-view feature extraction module will not be used.

\subsection{Distance Metrics}

During testing, we use DRA-OVG to obtain $f_o$, $f_g$, and $f_d$. For a vehicle, $f_f$ and $f_b$ come from $f_o$ and $f_g$ and represent the front view and back view features, respectively. The distance between images is calculated by:
\begin{equation}
\begin{aligned}
dist=
w_1\cdot \Big( \sqrt{(f_f^q-f_f^c)^2}+ \sqrt{(f_b^q-f_b^c)^2}\Big) \\+
w_2\cdot \sqrt{(f_d^q-f_d^c)^2},
\end{aligned}
\end{equation}
where $f_d^q$, $f_f^q$, and $f_b^q$ represent the discriminative, front view, and back view features of query images, respectively. $f_d^c$, $f_f^c$, and $f_b^c$ represent the discriminative, front view and back  view features of candidate images, respectively. $w_1$ and $w_2$ are hyper-parameters.
\section{Experiment}
In this section, we first introduce the datasets and show some implementation details of our method. Then some quantitative and qualitative experiments are shown to evaluate the effectiveness of all proposed components. Finally, we compare our method with some state-of-the-art methods on VehicleID \cite{liu2016deepl} and VeRi-776 \cite{liu2016deep} datasets.

\subsection{Dataset}
Experiments are mainly conducted on the general VehicleID dataset. Since each vehicle in the dataset has mainly two orthogonal views (front and back), the problems of inter-class similarity and intra-class differences are severe at the same time. Thus it can fully verify the effectiveness of the proposed method. In addition, there are three test sets in the dataset, which can effectively reduce the negative impact of random sampling to a certain extent, so it makes more sense to study on the VehicleID dataset. There are 221,763 images of 26,267 vehicles in the dataset captured by different surveillance cameras in the city. The training set contains 110178 images, and the test set contains 111585 images. Following the settings in \cite{liu2016deepl}, we use three test subsets of different sizes: 800, 1600, and 2400. 

The VeRi-776 dataset contains 776 vehicle IDs captured by 20 cameras, and each car has multiple viewpoints. The dataset contains 576 vehicles with 37,778 images for training and 200 vehicles with 11,579 images for testing. An additional set of 1,678 images selected from the testing set are used as query images. We strictly follow the evaluation protocol proposed in \cite{liu2016deep}.

\subsection{Implementation Details}
 The algorithm is implemented by Pytorch with the GPU mode and runs on the machine with Geforce GTX 1060 GPU, 6 GB memory, and i5 3470 CPU. In order to explore the mapping rules of vehicle semantic features in the feature space, we randomly sampled 1,000 vehicle images from the training set to construct the offline subset. We choose VGG-16 as the basic network $F$ to locate the semantic regions accurately due to its good position correspondence between features and semantics regions in original images (some other common-used networks, such as ResNet and DenseNet, have also been tested to extract features, but their positioning results of semantics are not satisfied as that by VGG-16). Thus, the structure of $F$ is the same as VGG-16 pre-trained on ImageNet. The weights of the $F$ network are shared throughout the algorithm. When we train it on vehicle datasets, the Adam algorithm and cross-entropy loss are used to optimize the network. Furthermore, the base learning rate is 0.001, which decreases by multiplying 0.1 after every 15 epochs. The batch_size is set to 64. Images are resized to 224 × 224 for both training and testing. The training of $F$ is stopped after 100 epochs, and then its parameters are fixed. The generation network is designed as a four-layer fully connected structure with 2048, 4096, 2048, and 1024 output dimensions. The normalization layer and activation layer are used between fully connected layers, and the activation function is 'ReLu'.The training of the generator net and the fusion net is stopped after 50 and 45 epochs, respectively. We use the SpectrumClustering algorithm in Sklearn to cluster features. Parameters are set as:  n_clusters=5, eigen_solver = None, assign_labels = 'k_means', random_state = None, and other parameters are default values. $\alpha_1$ and $\alpha_2$ are set to 0.1 and 0.9, respectively, and we only use vehicles' ID labels. The Cumulative Matching Characteristic (CMC) curve and the mean average precision (mAP) are used to evaluate our method.

\subsection{Qualitative results of viewpoint feature generation}

\begin{figure*}[htb]

	\center{
		\includegraphics[width=12cm]{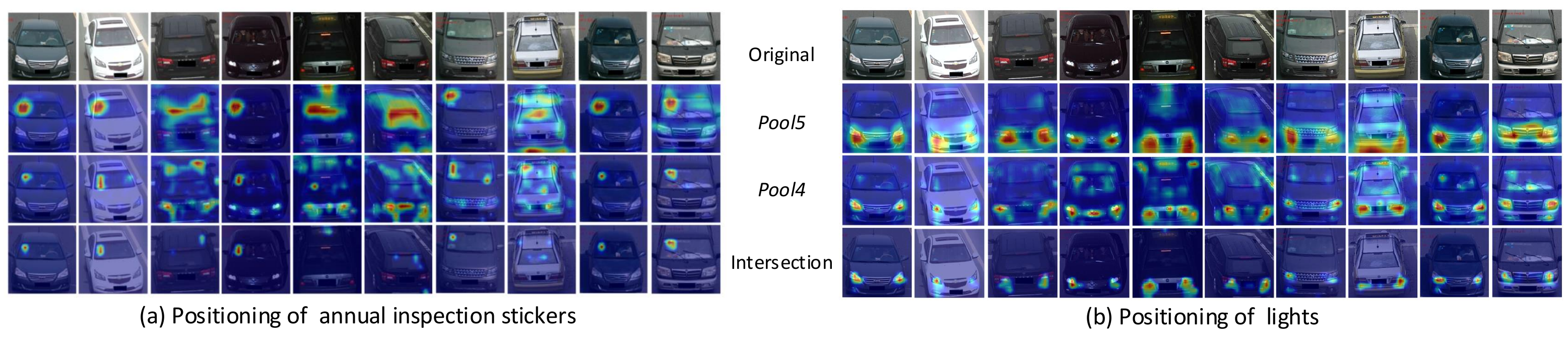}
\caption{The positioning results of annual inspection stickers and lights on VehicleID}
	\label{picture five} }
	\end{figure*}

Fig. \ref{picture five} shows some positioning results of annual inspection stickers and lights. We obtain satisfactory location results by combining the regions located by the feature maps of $pool4$ and $pool5$ layers. For annual inspection stickers, to eliminate the slight interference from the back of vehicles, we used 0.05 as the threshold for dividing the corresponding $P^r$ as two parts: 1 and 0. Moreover, the positioning of lights is always accurate. The results qualitatively demonstrate that our method can accurately locate the annual inspection stickers and lights of vehicles.
\subsection{Qualitative Results of Discriminative Region Location}

\begin{figure*}[htb]
	\center{
		\includegraphics[width=12cm]{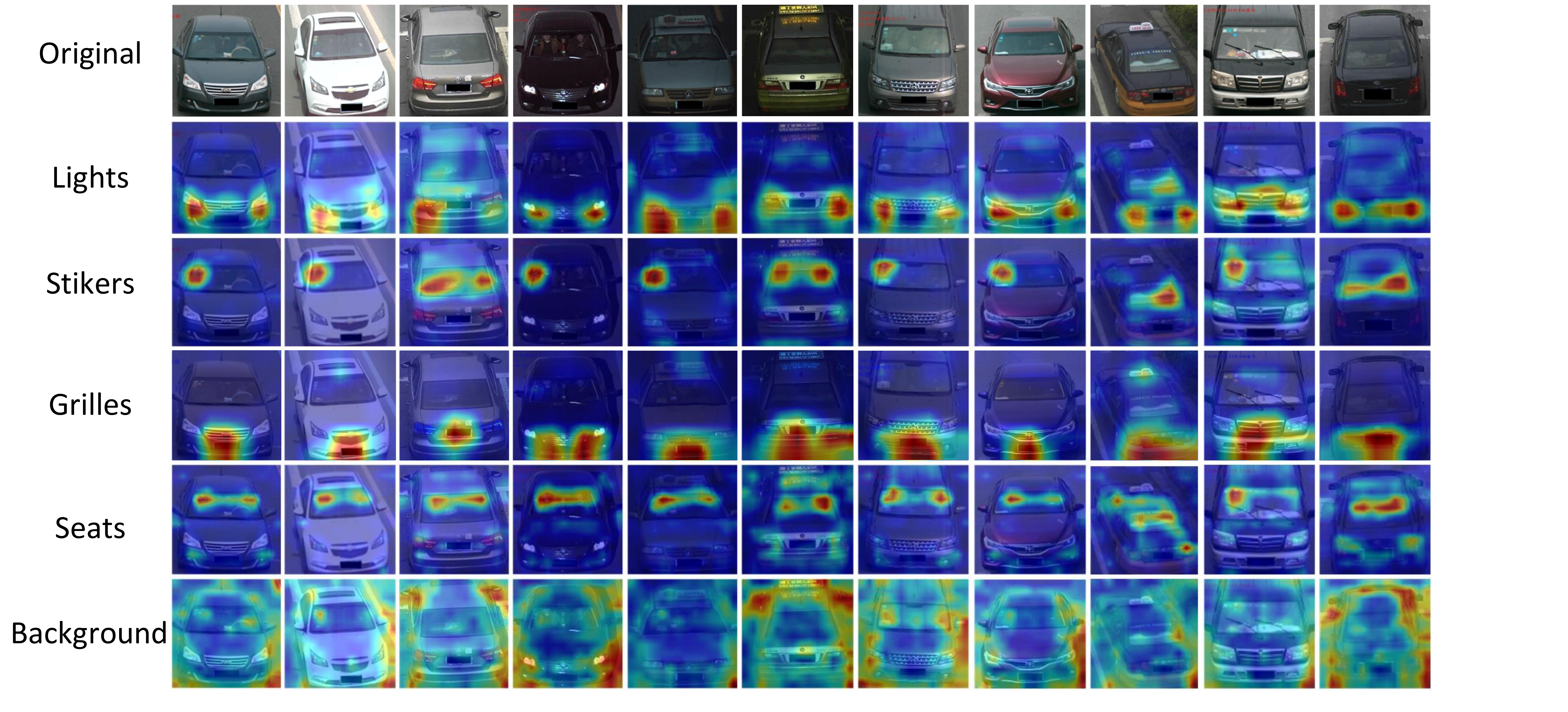}
		\caption{ Results of different semantic regions positioning. The first row is the original vehicle images, and the rest rows are the heat maps of different semantics located by semantic prototypes on VehicleID.
	}\label{picture six} }
\end{figure*}

Fig. \ref{picture six} shows some examples of the localization results of different semantic regions. It can be visually observed that the prototypes can automatically detect various parts of vehicles. After knowing what part of the vehicle each semantic prototype detects, we can select discriminative regions by ablation experiment shown in Table \ref{Table four}. There are two reasons for choosing the feature maps of the $pool5$ layer of the $F$. First, the deep convolution kernel has a sizeable sensory field. The features of each position in the feature map can include rich semantics while paying attention to detailed cues, hence avoiding some naive errors caused by narrow vision. As shown in Fig. \ref{picture six}, the regions of each semantic located by the semantic prototype also contain parts of the vehicle body. Secondly, the deeper convolution kernels represent more advanced and unique semantics, which can be accurately positioned. The third row of Fig. \ref{picture five} shows the location results, which are located by the feature maps of the $pool4$ layer. As we can see, although different semantic regions can be located, there are many mistake locations.

\subsection{Evaluation of Viewpoint Identification}

Before we optimize $G_f(\cdot)$, we must demonstrate the effectiveness of the viewpoint discriminator. We randomly selected 200, 500, and 1000 images of vehicles from the test set to evaluate the discriminators. Table \ref{Table two} shows the accuracy of the viewpoint judgment based on the features of different layers.

\begin{figure}[htb]
	\center{
		\includegraphics[width=10cm]{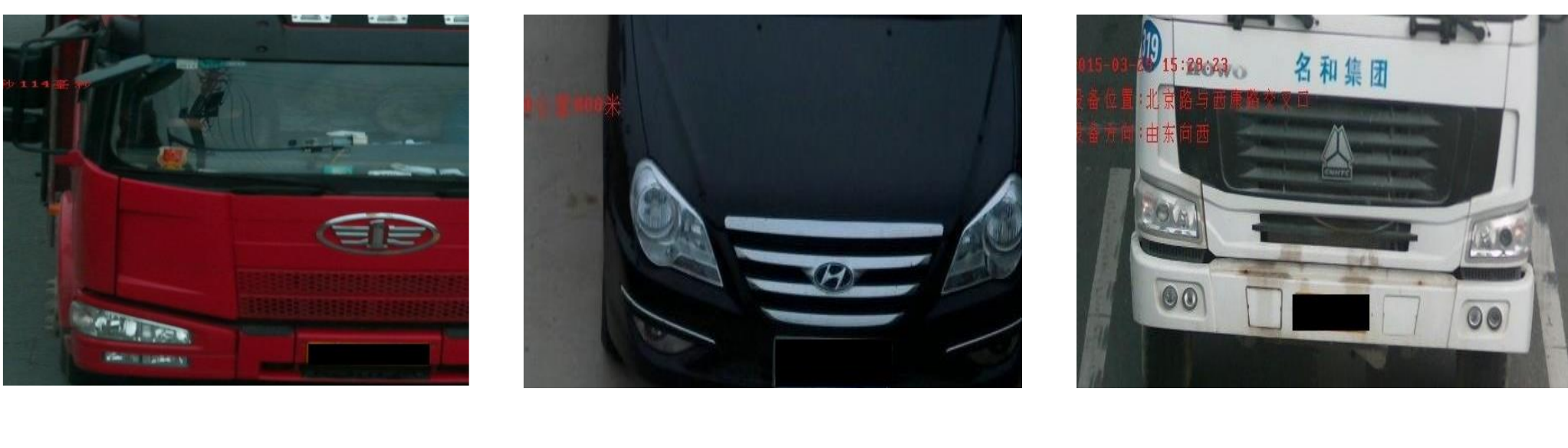}
		\caption{Viewpoint identification error cases display on VehicleID.
		}\label{picture eleven} }
\end{figure}

As shown in the last row of Table \ref{Table two}, under the three test sets, all the accuracies of the viewpoint identification exceed 95\%. In order to analyze the algorithm more deeply, we output some cases where the viewpoint identification is wrong. As shown in Fig. \ref{picture eleven}, the images that cannot be correctly judged all contain only part of the vehicle body. Since these images lack the features of the annual inspection stickers or lights, the algorithm fails in these situations. Excluding these extreme cases, the viewpoint identification algorithm hardly makes mistakes.
\begin{table}[]
	\centering
	\begin{center}
	\caption{Evaluation ($\%$) of viewpoint identification accuracy on VehicleID.} \label{Table two}
	\begin{tabular}{c c c c}
		\hline\hline
		Layers & Size=200 & Size=500 & Size=1000 \\ \hline
		$pool4$    & 50.9     & 53.4     & 55.9      \\
		$pool5$    & 62.3     & 55.7     & 58.2      \\
		$pool4$-$pool5$ & 99.0     & 97.4     & 96.5      \\ \hline\hline
	\end{tabular}
	\end{center}
\end{table}
The quantitative results in Table \ref{Table two} suggest that our vehicle viewpoint identification method can provide reliable supervisory information for training $G_f(\cdot)$. It can also be deduced from the data that using feature maps of a single layer is not enough to distinguish viewpoints accurately.

As shown in Table \ref{Table nine}, in order to select the most appropriate number of clusters for our method, the feature set $X_{set}$ is clustered into different numbers of groups. As can be seen from Table \ref{Table nine}, when c = 5, DRA achieves the best performance, so we set the number of clusters to 5 in this paper. 

As shown in Table \ref{Table three}, we try to cluster viewpoint features of the images in the subset into different numbers of clusters so that we can choose the most suitable number of generated views for the OVG model. We can see that when $ k $ = 2, the performance of vehicle Re-ID is the best. That is mainly due to the following reasons: First, the identity features of the vehicle are mainly concentrated on some specific views (such as the front view and back view), so generating more views will not introduce more practical information. Secondly, because the features of the generated views are just a guess, the original view should play a major role in identifying vehicles. If we generate more views, the feature of the original view will have less influence on the distance measurement, which is harmful to vehicle Re-ID. Finally, since there is a certain imbalance in the vehicle datasets, not every vehicle has images from so many different viewpoints. Those vehicles that lack images under specific viewpoints have to be discarded, reducing the number of images in the training set. That undoubtedly harms network training. Therefore, we finally choose to generate two views.
\begin{table}[]
	\centering
	\begin{center}
	\caption{Evaluation (mAP$\%$) of the DRA model on Veri-776 and VehicleID (size=2400). $c$ is the cluster number of features.} \label{Table nine}
	\begin{tabular}{c c c c c c c}
		\hline\hline
		Clusters & Veri-776  & VehicleID  \\   \hline
		$c$=4    & 74.41     & 79.25      \\
		$c$=5    & 75.01     & 80.71      \\
		$c$=6    & 74.73     & 79.38      \\
		$c$=7    & 74.56     & 78.11      \\
		$c$=8    & 73.68     & 76.34      \\\hline\hline
	\end{tabular}
	\end{center}
\end{table}

\begin{table}[]
	\centering
	\begin{center}
	\caption{Evaluation ($\%$) of the OVG model. $k$ is the cluster number of the viewpoint features on VehicleID.} \label{Table three}
	\begin{tabular}{c c c c}
		\hline\hline
		Clusters & mAP & r=1 & r=5 \\ \hline
		$k$=2    & 56.47     & 55.12     & 63.24 \\
		$k$=3    & 47.39     & 45.50     & 52.75 \\
		$k$=4    & 42.24     & 40.04     & 48.58 \\ \hline\hline
	\end{tabular}
	\end{center}
\end{table}

\subsection{Ablation Studies}

\begin{table}[]
\renewcommand\tabcolsep{3.5 pt}
	\begin{center}
	\centering
	\caption{Evaluation($\%$) of the impact of different semantics on the accuracy of Re-ID on VehicleID (size=2400).}\label{Table four}
	{
	\begin{tabular}{ccccc|ccc}
		\hline\hline
		light & sticker & grille & seat & background         & mAP  & r=1 & r=5   \\ \hline
		\checkmark  & & & &             & 44.93  & 41.17  & 50.24 \\
		\checkmark  & \checkmark& & &        & 68.25 & 66.91 & 78.17 \\
		\checkmark  & \checkmark& \checkmark& &                 & 80.71 & 78.80 & 90.00 \\
		\checkmark  & \checkmark& \checkmark& \checkmark&          & 80.07 & 78.53 & 89.41 \\ 
		\checkmark  & \checkmark& \checkmark& \checkmark& \checkmark                     & 78.47 & 76.03 & 87.33 \\ \hline\hline
	\end{tabular}}
	\end{center}

\end{table}
As shown in Table \ref{Table four}, we use the features of different semantics to identify vehicles. It can be seen that the recognition accuracy is the highest when we use the features of lights, air intake grilles, and annual inspection stickers. The introduction of the features of seats and background reduces the recognition accuracy. That is because the passengers on seats and the background of images vary among images. The features of these two semantics cannot provide a reliable basis for vehicle Re-ID. Thus we finally choose the regions of light, air intake grille, and annual inspection sticker as the discriminative regions.

As shown in the first part of Table \ref{Table five}, the Global-feat is the feature obtained by conducting global average pooling on the output of the pool5 layer. We feed the Global-feat into the fusion net to conduct classification training, and then the Global+$\mathcal{L}_{Reid}$ is obtained. As shown in Fig. \ref{picture two}(b), $\mathcal{L}_{Reid}$ is the weighted sum of cross-entropy loss and triplet loss. As with Global-feat, DRA-feat is the feature obtained by conducting global average pooling on the discriminative region features.

\begin{table}[]
	\begin{center}
	\centering
	\caption{Evalution ($\%$) of effectiveness of the DRA-OVG on VehicleID (size=2400).}\label{Table five}
	{
	\begin{tabular}{ccccccc}
		\hline\hline
		Baselines       &        & mAP  & & r=1  & & r=5   \\ \hline
		Global-feat     &       & 76.26 & & 74.66 & & 86.45 \\
		Global+$\mathcal{L}_{Reid}$  &     & 78.62 & & 76.13 & & 87.51 \\
		DRA-feat       &         & 72.53 & & 70.07 & & 84.54 \\
		DRA            &         & 80.71 & & 78.80 & & 90.00 \\ \hline
		OVG            &         & 56.47 & & 55.12 & & 63.24 \\
		OVG-Global+$\mathcal{L}_{Reid}$  & & 81.88 & & 79.11 & & 91.37 \\ \hline
		DRA-OVG        &         & 82.51 & & 79.25 & & 91.03 \\ \hline\hline
	\end{tabular}}
	\end{center}

\end{table}

\begin{figure}[htb]
	\center{
		\includegraphics[width=8.5cm]{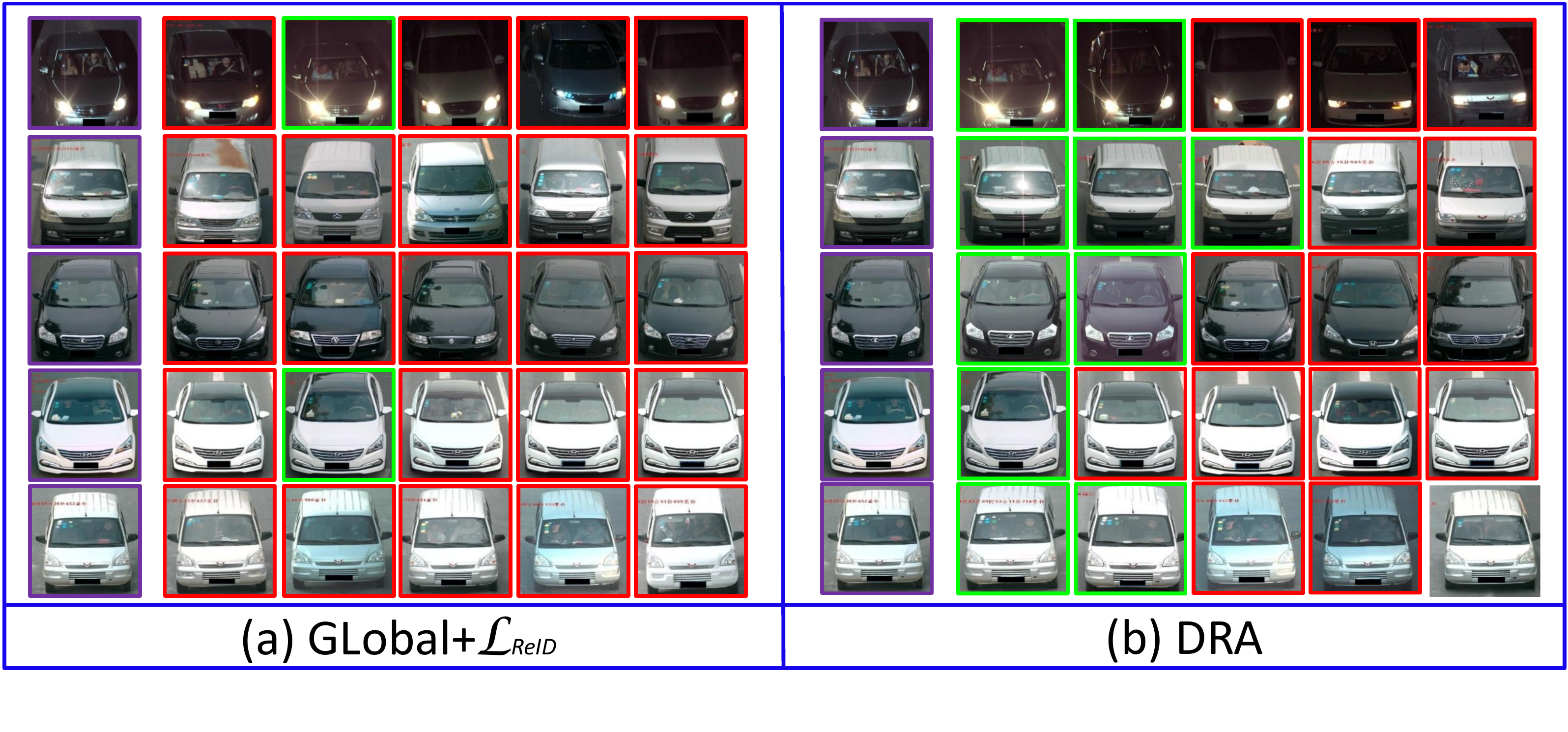}
		\caption{Comparisons of qualitative results based on global features and discriminative features on VehicleID. Query vehicles are framed in purple boxes, red boxes and green boxes frame the wrong and the correct results respectively .
	} \label{picture seven}}
\end{figure}

Part one of Table \ref{Table five} shows that, since DRA eliminates interference from unrelated regions, it achieves better generalization performance than Global+$\mathcal{L}_{Reid}$ through the same Re-ID training. Compared with the Global+$\mathcal{L}_{Reid}$, the DRA model increases the mAP by 2.09$\%$ and increases the rank1 by 2.67$\%$. These improvements suggest that the proposed DRA model is helpful to distinguish similar-looking vehicles. Furthermore, some retrieval results, which are shown in Fig. \ref{picture seven}, qualitatively illustrate that most gallery candidates retrieved by the global feature have similar irrelevant details, such as the same passengers, sunlight-reflection points, and landscape reflections. Our DRA model can eliminate the influence of those irrelevant features and rank the candidates with similar discriminative region features at the top position.

\begin{figure}[htb]
	\center{
		\includegraphics[width=8.5cm]{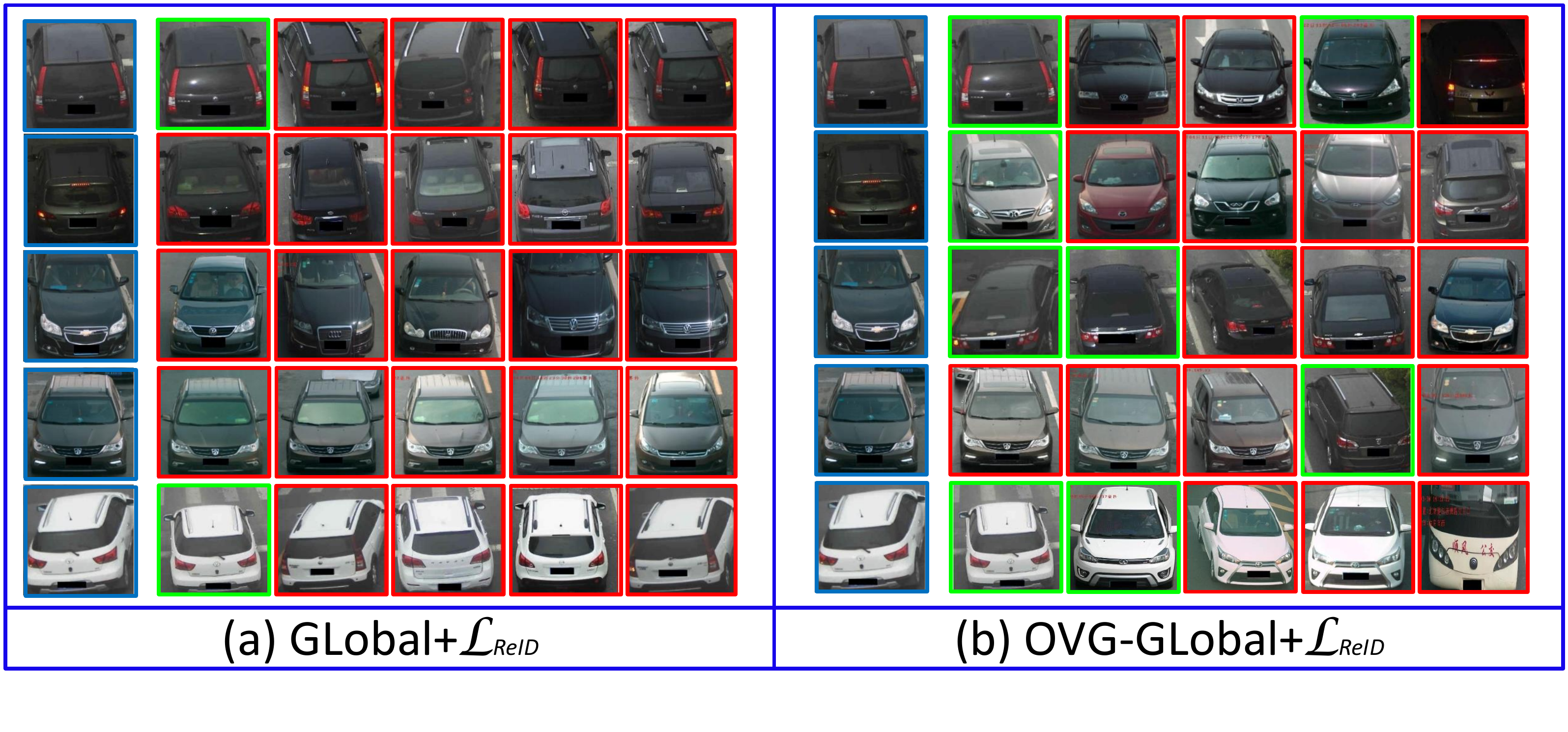}
		\caption{Comparisons of qualitative results based on global features and OVG-Global+$\mathcal{L}_{Reid}$ on VehicleID. Query vehicles are framed in purple boxes, red boxes and green boxes frame the wrong and correct results respectively.
		}\label{picture eight} }
\end{figure}

\begin{figure}[htb]
	\center{
		\includegraphics[width=7cm]{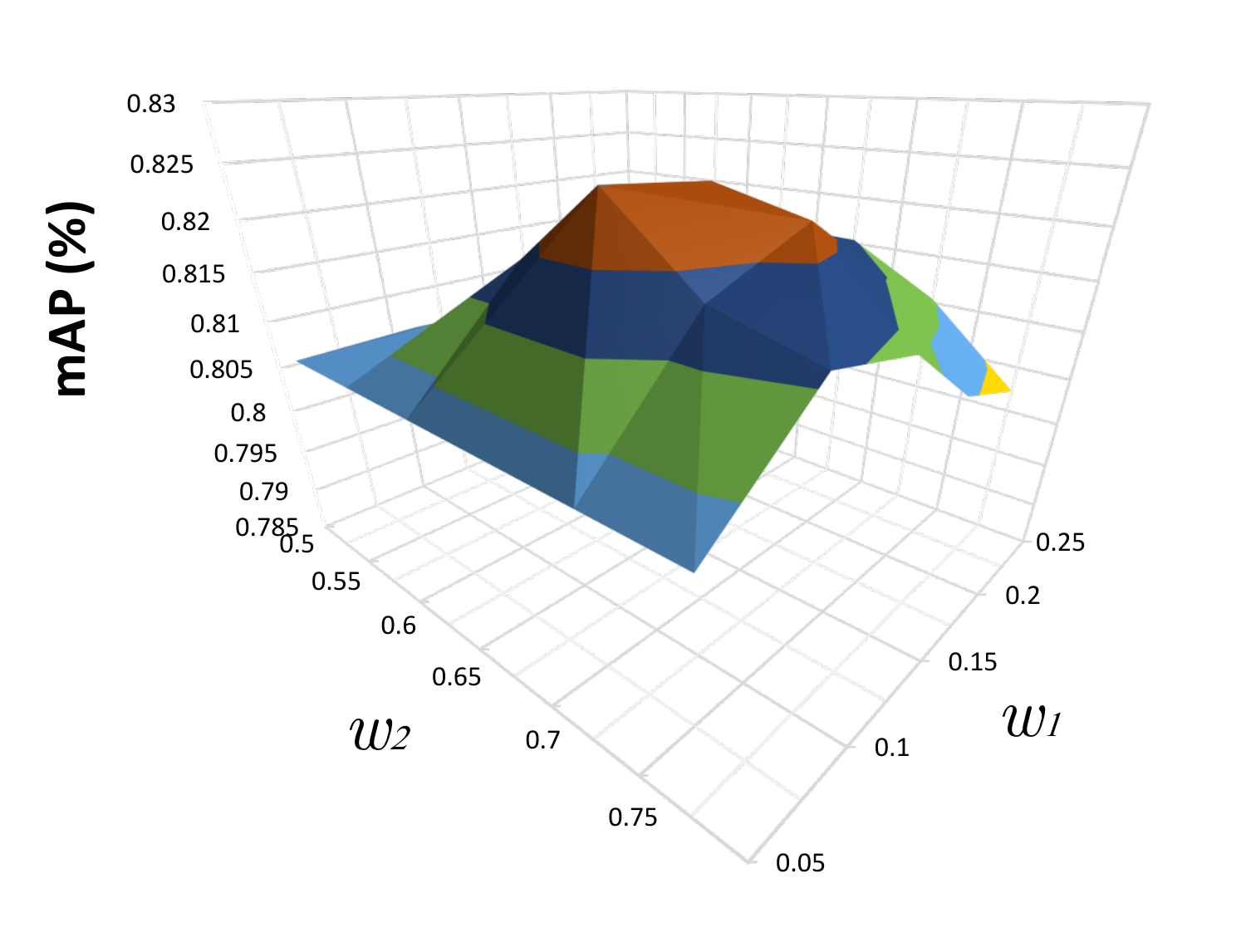}
		\caption{The mAP surface of DRA-OVG on VehicleID (size=2400) when changing the value of $w_1$ and $w_2$ in (7).
		}\label{picture ten} }
\end{figure}

As shown in the second part of Table \ref{Table five}, we cannot achieve excellent Re-ID performance using only the OVG model because the generated orthogonal-view features are only a guess easily disturbing by similar vehicles. The OVG model is more suitable as a supplement to other models rather than a substitute. The mAP and the rank1 are increased by 3.26$\%$ and 2.98$\%$, respectively, when combining OVG with the Global+$\mathcal{L}_{Reid}$. These meaningful improvements indicate that the proposed OVG model is undoubtedly profitable to vehicle Re-ID. Moreover, Fig. \ref{picture eight} qualitatively demonstrates the effectiveness of the OVG model. Most gallery candidates retrieved by global features have the same viewpoint as the query image. On the contrary, our OVG model can propose more gallery candidates that are under different viewpoints.
The last row of Table \ref{Table five} shows that our DRA-OVG model performs best. Compared with the Global+$\mathcal{L}_{Reid}$, our model improves the mAP value by 3.89$\%$ and the rank1 value by 3.12$\%$. As shown in Fig. \ref{picture ten}, the optimal ratio of $w_1$ and $w_2$ is 0.1 to 0.65.

In addition to evaluating the accuracy of Re-ID, we also test the speed of our method: Extracting discriminative features for each image takes 80 ms (about 12 fps), and extracting the viewpoint feature for an image takes 76 ms (about 13 fps). Performing a distance measurement between two images takes about 90 ms (about 11 fps).
\begin{table*}[]
	\begin{center}
	\centering
	\caption{Comparisons with state-of-the-art Re-ID methods on VehicleID.}\label{Table six}
	\resizebox{12cm}{2.5cm}
	{
	\begin{tabular}{cc cccc cccc ccc}
		\hline\hline
		                      & & \multicolumn{3}{c}{Test size=800}  & & \multicolumn{3}{c}{Test size=1600}&&\multicolumn{3}{c}{Test size=2400} \\ \cline{3-5} \cline{7-9} \cline{11-13}
		 Methods            & Backbones                    &mAP     & r=1      & r=5   & &mAP  & r=1   & r=5   & &mAP& r=1             & r=5    \\ \hline
		LOMO\cite{liao2015person} & -                      &-       & 19.76    & 32.01 & &-    & 18.85 & 29.18 & &-    &15.32          & 25.29  \\
		GoogLeNet\cite{yang2015large} & GoogLeNet          &-       & 47.88    & 67.18 & &-    & 43.40 & 63.68 & &-    &38.27          & 59.39  \\
	    FACT\cite{liu2016deep}  &	GoogLeNet              &-       & 49.53    & 68.07 & &-    & 44.59 & 64.57 & &-    &39.92          & 60.32  \\
		DAVR\cite{peng2020cross} & -                       &54.01   & 49.48    & 68.66 & &49.72& 45.18 & 63.99 & &45.18&40.71          & 59.02   \\
		Mixed Diff+CCL\cite{liu2016deepl} & VGGM           &54.60   & 48.90    & 73.50 & &48.10& 42.80 & 66.80 & &45.50&38.20          & 61.60  \\
		EALN\cite{lou2019embedding} & VGGM                 &77.50   & 75.11    & 88.09 & &74.20& 71.78 & 83.94 & &71.00&69.30          & 81.42   \\
		GS-TRE loss \cite{bai2018group} &VGGM              &75.40   & 75.90    & 84.20 & &74.30& 74.80 & 83.60 & &72.40&74.00          & 82.70  \\ \hline
		XVGAN\cite{zhou2017cross} & -                      &-       & 52.87    & 75.65 & &-    & 49.55 & 68.85 & &-    &44.89          & 63.38  \\
	    VAMI\cite{zhou2018aware} & -                       &-       & 63.12    & 80.83 & &-    & 52.87 & 71.79 & &-    &47.73          & 66.65  \\\hline
		AGNet-ASL-LD\cite{wang2020attribute} & -           &74.05   & 71.15    & 83.78 & &69.23& 69.23 & 81.41 & &69.66&65.74          & 78.28   \\	
		TAMR \cite{guo2019two} & ResNet-18                 &67.64   & 66.02    & 79.71 & &63.69& 62.90 & 76.80 & &60.97&59.69          &73.87   \\
		AAVER\cite{khorramshahi2019dual} &ResNet-101       &-       & 74.69    & 93.82 & &-    & 68.62 & 89.95 & &-    &63.54          & 85.64   \\
		PDFP\cite{he2019part} & ResNet-50                  &-       & 78.40    & 92.30 & &-    & 75.00 & 88.30 & &-    &74.20          & 88.40   \\ \hline
		DRA(ours)  & VGG-16                                &84.51   & 82.13    & 93.43 & &83.04& 80.98 & 92.04 & &80.71&78.80          & 90.00     \\
		OVG-Global+$\mathcal{L}_{Reid}$(ours)& VGG-16      &84.43   & 82.01    & 93.47 & &82.16& 80.73 & 91.87 & &81.88&79.11          & \textbf{91.37}    \\ 
		DRA-OVG(\textbf{ours})    & VGG-16                 &\textbf{85.14}& \textbf{83.40}  & \textbf{94.02} & &\textbf{83.88}& \textbf{82.31}          & \textbf{93.20} & &\textbf{82.51}& \textbf{79.25}          & 91.03  \\ \hline\hline
	\end{tabular}}
	\end{center}
\end{table*}

\begin{figure*}[htb]
	\center{
		\includegraphics[width=8cm]{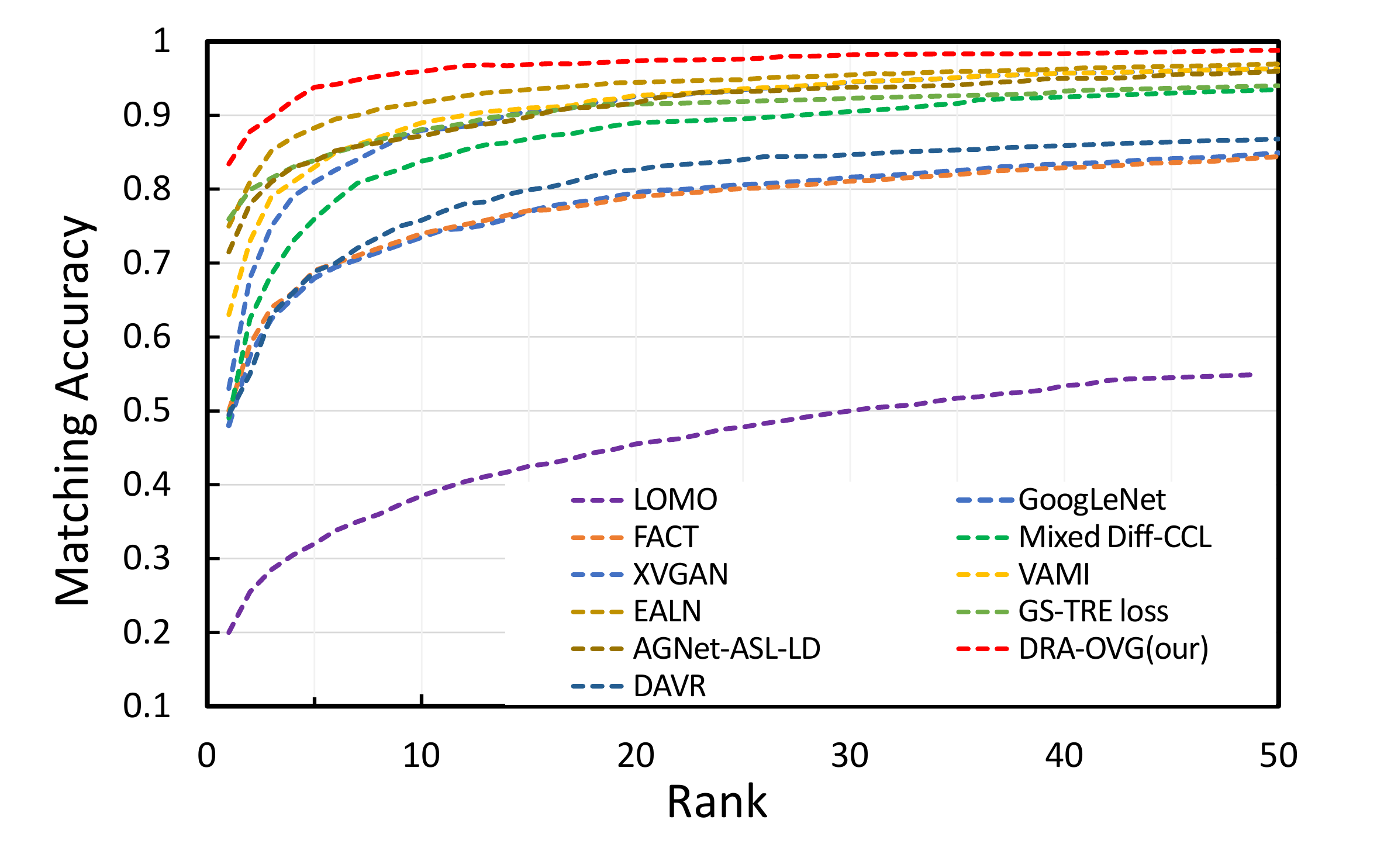}
		\caption{CMC curves of different Re-ID methods on VehicleID (size=800).
		}\label{picture nine} }
\end{figure*}
\subsection{Comparisons with State-of-the-Arts}
\subsubsection{On VehicleID}
On the VehicleID dataset, we compare our DRA-OVG with some state-of-the-art methods. The second part of Table \ref{Table six} shows the supervised methods based on multi-view inference. They use the features of the original views and the inferred views of vehicles simultaneously. Compared with VAMI\cite{zhou2018aware}, the OVG-Global+$\mathcal{L}_{Reid}$ increases by 31.38$\%$ on rank1 and by 24.72$\%$ on rank5 on the test size of 2,400. The third part of Table \ref{Table six} shows the supervised methods based on detailed cues. Among them, TAMR\cite{guo2019two} also extracts discriminative features from the visual appearance of vehicles. Compared with it, our DRA model increases by 19.11 $\%$ on rank1 and by 16.13$\%$ on rank5. Besides, PDFP\cite{he2019part} combines global features with local features to address the problem of vehicle Re-ID, and it also uses annual inspection stickers, lights, and air intake grilles as discriminative regions. Compared with it, our DRA model increases by 4.6 $\%$ on rank1 and by 1.6$\%$ on rank5 on the large size of the gallery. AAVER\cite{khorramshahi2019dual} is a two-path adaptive attention model for vehicle Re-ID. The method uses the manually marked key points to train the network to obtain the ability of discriminative region localization and viewpoint estimation. Compared with it, our DRA-OVG model increases by 15.71 $\%$ on rank1 and by 5.39 $\%$ on rank5 on the large size of the gallery. As shown in Fig. \ref{picture nine}, the CMC curves show the performance of methods more intuitively. As shown in Table \ref{Table six}, the proposed DRA-OVG model achieves better results than all existing approaches under the three settings of the gallery size. Therefore, we can conclude that our model is very beneficial to the vehicle Re-ID task.
\begin{table}[]
	\caption{Comparisons with state-of-the-art Re-ID methods on VeRi-776.}
		\label{Table seven}
	\centering
	\begin{center}
	{
		\begin{tabular}{cc c c c }
			\hline\hline
			Methods     & Backbones                            &mAP               & r=1             & r=5              \\ \hline
			LOMO\cite{liao2015person} & -                      & 7.98             & 23.87           & 39.14            \\
			GoogLeNet\cite{yang2015large}  & GoogLeNet         & 17.81            & 52.12           & 66.79            \\
			FACT\cite{liu2016deep}  & GoogLeNet                & 18.73            & 51.85           & 67.16            \\
			DAVR\cite{peng2020cross}   & -                     & 26.35            & 62.21           &73.66             \\ 
			GS-TRE loss \cite{bai2018group} & VGGM             & 59.47            & 96.24  &98.97             \\ \hline
			XVGAN\cite{zhou2017cross}  & -                     & 24.65            & 60.20           & 77.03            \\
			VAMI\cite{zhou2018aware}   & -                     & 50.13            & 77.03           & 90.82            \\\hline
			AGNet-ASL-ID\cite{wang2020attribute} & -           & 66.32            & 90.90           &96.20                  \\ 
			AAVER\cite{khorramshahi2019dual} & ResNet-101      & 61.18            & 88.97           &94.70                  \\
			PDFP\cite{he2019part}  & ResNet-50                 & 74.30            & 94.30           &98.70                   \\ 
			PCRNet\cite{liu2020beyond} & -                         & 78.60            & 95.40           &98.40\\ 
			PVEN\cite{meng2020parsing} & -                     & 79.50            & 95.60           &98.40\\ 
			VehicleNet\cite{zheng2020vehiclenet} & ResNet-50   & \textbf{83.41}   & \textbf{96.78}  &-\\ \hline
			Global+$\mathcal{L}_{Reid}$  & VGG-16              & 71.68            & 91.18           & 97.44                \\ 
			DRA(\textbf{ours})  & VGG-16                       & 75.01   & 94.01           & \textbf{98.98}           \\ \hline\hline
		\end{tabular}}
	\end{center}
\end{table}
\subsubsection{On VeRi-776}
On VeRi-776 dataset, we compare the DRA model with some state-of-the-art methods. As shown in Table  \ref{Table seven}, compared with the Global+$\mathcal{L}_{Reid}$, our method increases by 2.83 $\%$ on rank1 and by 1.54$\%$ on rank5. This result shows that, by eliminating the interference of irrelevant features, the DRA model can help the base net achieve a significant improvement on the VeRi-776 dataset. Besides, compared with some supervised methods based on detail cues in the third part of Table \ref{Table seven}, our method still has certain advantages. DRA increases by 0.71 $\%$ on mAP and by 0.28 $\%$ on rank5 when compared with PDFP\cite{he2019part}, which also uses annual inspection stickers, lights, and air intake grilles as discriminative regions.

Our method does lag behind some supervised methods. For example, VehicleNet\cite{zheng2020vehiclenet} uses extensive additional data from other datasets to enhance its recognition capability. PCRNet\cite{liu2020beyond} relies on a new large-scale Multi-grained vehicle dataset to learn discriminative part-level features, and PVEN\cite{meng2020parsing} annotates a subset of VeRi-776 for training vehicle part parsing network. Nevertheless, our method only uses ID labels and is trained under a single dataset. Thus the training cost of our method is relatively low than those methods. It can be seen from Table \ref{Table seven} that our DRA model brings apparent benefits to the vehicle Re-ID on VeRi-776 using only the ID label. 

Moreover, due to the unsupervised limitation, our method can only partially solve the cross-view recognition problem (between the front and back views) for the time being. For multi-view recognition and inference, our OVG method is still slightly insufficient. Therefore, under the VeRi-776 dataset, the OVG model is not very helpful for improving the performance of vehicle Re-ID, which is worthy of further study.

\section{Discussion}

\begin{figure}[htb]
	\center{
		\includegraphics[width=8.5cm]{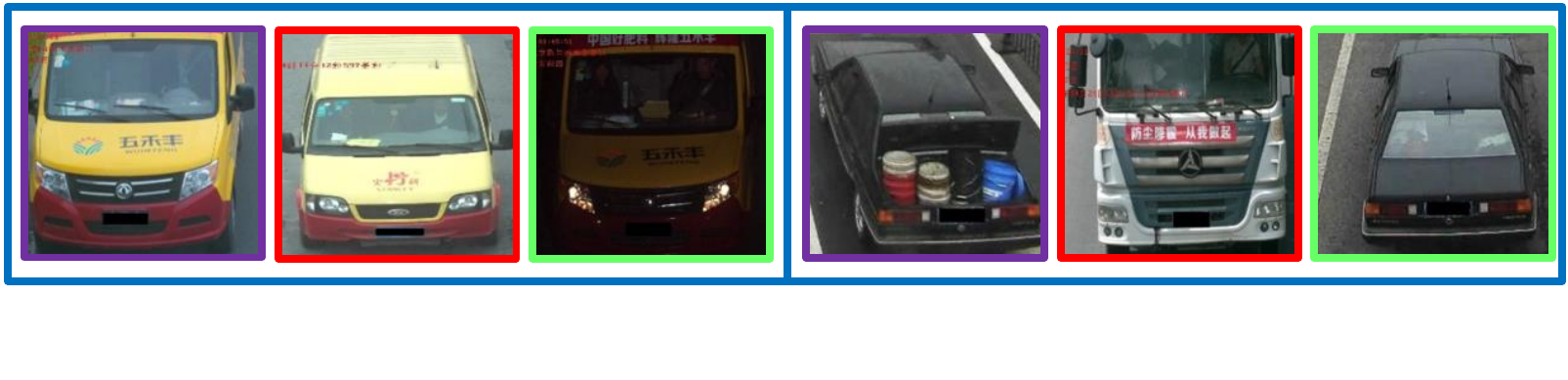}
		\caption{Some hard retrieval samples. Query vehicles are framed in purple boxes, red boxes and green boxes frame the wrong and correct results respectively
		}\label{picture twelve} }
\end{figure}

As shown in Fig. \ref{picture twelve}, there are some hard retrieval samples. For the first case, the identification error is caused by the great difference in illumination. For the second case, the identification error is caused by some excellent appearance change of the vehicle. These extreme conditions seriously affect extracting the features of discriminative regions and the inference of orthogonal features, so our method cannot successfully identify them yet.

For the optimal number of clusters of $X_{set}$, we determine it as five according to the recognition accuracy of the DRA model on the two datasets used in this paper. However, it does not mean that five is always the optimal setting for other datasets. It is likely to be data-dependent or model-dependent.

The proposed semantic localization method and the viewpoint identification method can be considered almost free of computation because we only need to generate semantic prototypes and viewpoint discriminators once offline and perform the vector distance measurement operation in the online learning stage. These two processes need very little calculation. In addition, before training the fusion net $T(\cdot)$, we can use $F$ and the semantic prototypes to extract discriminative region features of all the images in the training set and save them. These features can be directly used during the training of the fusion net $T(\cdot)$, which significantly speeds up the training. Similarly, for the training of the generation network $G(\cdot)$, we can also generate the $f_o$, $\hat{f_g}$ and the viewpoint tag for each image in advance. In the online learning stage, $F$ is not required, so the training cost of our method is meager.
 
Vehicle Re-ID is a fine-grained image categorization problem. In this problem, samples often have similar global appearances, and the differences only exist in local semantics. Therefore, introducing attention mechanisms in categorization algorithms is a common method to address such problems. The positioning of local semantic regions crucial for algorithms based on the attention mechanism. So the proposed unsupervised positioning method of semantic regions may have some inspirations for other fine-grained categorization problems.

This paper mainly divided vehicle viewpoints into two categories (the front and back). Because the side view of vehicles lacks the annual inspection sticker feature, its viewpoint feature is closer to the back discriminator in the feature space than the front one. Therefore, given an image taken from the side of a vehicle, the OVG model will treat it as an image from the back viewpoint and use its features to generate the features of the vehicle's front view.

The OVG model proposed in this paper is mainly dedicated to generating the features of vehicles' front and back views to help address the vehicle Re-ID problem. That is due to the following two reasons: First, the front and back views of a vehicle are very different, and they contain a wealth of detailed information that can be used to represent the vehicle's identity. So the combination of the front and back features is of great help in re-identifying vehicles. However, the two flanks are symmetrical and similar, and only a few features can be used to identify the vehicle's identity. Therefore, generating vehicles' flank features is relatively less helpful for vehicle Re-ID. Second, our viewpoint identification algorithm is based solely on the overall data distribution due to the unsupervised restriction. There are certain limitations in identifying more refined vehicle viewpoints.

\section{Conclusions}

In this paper, we tried to conquer the multiple challenges encountered in vehicle Re-ID tasks by only using the ID label of vehicles. We proposed a method leveraging the mapping consistency of vehicles' local semantics in feature space to generate semantic prototypes. Moreover, these prototypes were used to filter and transform different semantic features directly in feature space. Furthermore, we designed a DRA-OVG model, which could extract and fuse the discriminative regional features of vehicles and generate their orthogonal-view features based on the input view features. Finally, vehicles' identity features were constructed in the complete feature space to optimize pairwise distances.

Extensive experiments were conducted to verify the efficiency of the proposed method. First, the heat maps suggested that the prototypes could accurately locate various semantic regions in a vehicle image. Second, the ablation experiments verified the effectiveness of the ingredients in the model. Third, the DRA-OVG model is compared with the state-of-the-art methods on VehicleID and VeRi-776 datasets. The results reveal that our method could still surpass most advanced methods with only vehicles' ID labels.

Although the investigation of this paper may have some inspirations for unsupervised viewpoint identification and multi-view feature generation, how to use a vehicle's flank view under unsupervised conditions to construct the vehicle's identity features and generate multi-view features is still a task worth studying.

\section*{Declarations}
\begin{itemize} 
\item Funding:

This work is supported by the National Nature Science Foundation of China (grant No.61871106 and No.61370152), Key R \& D projects of Liaoning Province, China (grant No. 2020JH2/10100029), and the Open Project Program Foundation of the Key Laboratory of Opto-Electronics Information Processing, Chinese Academy of Sciences (OEIP-O-202002).

\item Conflict of interest/Competing interests:

All authors certify that they have no affiliations with or involvement in any organization or entity with any financial interest or non-financial interest in the subject matter or materials discussed in this manuscript.
\end{itemize}

\bibliography{bibliography}


\end{document}